\documentclass{article}

\usepackage{amsmath,amsfonts,bm}

\def\eqref#1{equation~\ref{#1}}
\def\Eqref#1{Equation~\ref{#1}}

\def\1{\bm{1}}

\def\vzero{{\bm{0}}}

\def\vmu{{\bm{\mu}}}
\def\vtheta{{\bm{\theta}}}

\def\vk{{\bm{k}}}

\def\vo{{\bm{o}}}

\def\vu{{\bm{u}}}

\def\vx{{\bm{x}}}
\def\vy{{\bm{y}}}
\def\vz{{\bm{z}}}

\def\mK{{\bm{K}}}

\def\mQ{{\bm{Q}}}
\def\mR{{\bm{R}}}

\def\mU{{\bm{U}}}

\def\mX{{\bm{X}}}
\def\mY{{\bm{Y}}}
\def\mZ{{\bm{Z}}}

\def\mSigma{{\bm{\Sigma}}}

\DeclareMathAlphabet{\mathsfit}{\encodingdefault}{\sfdefault}{m}{sl}
\SetMathAlphabet{\mathsfit}{bold}{\encodingdefault}{\sfdefault}{bx}{n}

\newcommand{\E}{\mathbb{E}}

\newcommand{\KL}{D_{\mathrm{KL}}}

\usepackage{graphicx}

\usepackage[utf8]{inputenc}
\usepackage{mathtools}
\usepackage{nccmath}
\usepackage{amsmath}
\usepackage{amssymb}
\usepackage{amsfonts}
\usepackage{xcolor}
\usepackage{xspace}

\usepackage{tikz}
\usetikzlibrary{arrows}
\usetikzlibrary{fit}
\usepackage{pgfplots}
\pgfplotsset{compat=newest}
\usepgfplotslibrary{groupplots}
\usepgfplotslibrary{dateplot}

\definecolor{bittersweet}{rgb}{1.0, 0.44, 0.37}

\newcommand{\tran}{^\intercal}
\newcommand{\minus}{\text{-}}
\newcommand{\inv}{^{\text{-}1}}
\newcommand*{\N}{\mathcal{N}}
\newcommand*{\vphi}{{\bm{\phi}}}

\newcommand*{\vxi}{{\bm{\xi}}}
\newcommand*{\vpsi}{{\bm{\psi}}}

\renewcommand*{\E}{\mathbb{E}}
\renewcommand*{\KL}{\mathbb{D}_{\text{KL}}}

\newcommand{\itwoc}{\textsc{i2c}\xspace}

\begin{document}
\title{Active Inference or Control as Inference?\\
	   \Large A Unifying View
  }
\author{
~
Abraham Imohiosen*$^\dagger$
\and
Joe Watson*$^\text{\textsection}$
\and
Jan Peters$^\text{\textsection}$
}

\date{
$^\dagger$RWTH Aachen University, Germany\\
$^\text{\textsection}$IAS, Technical University Darmstadt, Germany\\
\texttt{abraham.imohiosen@rwth-aachen.de}\\
\texttt{\{watson,peters\}@ias.informatik.tu-darmstadt.de}
}
\maketitle              %
\begin{abstract}
Active inference (AI) is a persuasive theoretical framework from computational neuroscience that seeks to describe action and perception as inference-based computation.
However, this framework has yet to provide practical sensorimotor control algorithms that are competitive with alternative approaches.
In this work, we frame active inference through the lens of control as inference (CaI), a body of work that presents trajectory optimization as inference.
From the wider view of `probabilistic numerics', CaI offers principled, numerically robust optimal control solvers that provide uncertainty quantification, and can scale to nonlinear problems with approximate inference.
We show that AI may be framed as partially-observed CaI when the cost function is defined specifically in the observation states. 

\end{abstract}
\section{Introduction}
Active inference (AI) \cite{10.3389/fnbot.2018.00045,friston2010free,friston2017active} is a probabilistic framework for sensorimotor behavior that enjoyed sustained interest from computational neuroscientists.
However, its formulation has been criticized for its opacity and similarity to optimal control \cite{gershman2019does,tweet,doi:10.1080/17588928.2015.1051952}, but is seemingly difficult to translate into an equally effective algorithmic form.
In this work, we offer a critical analysis of AI from the view of control as inference (CaI) \cite{Attias03planningby,Kappen_2005,levine2018reinforcement,todorov2007linearly,06-toussaint-ICML,i2corl}, the synthesis of optimal control and approximate inference.
The goal is to appreciate the insights from the AI literature, but in a form with computational and theoretical clarity.

\section{Background}
Here we outline the foundational theory and assumptions in this work.
\subsection{Problem Formulation}
We specifically consider a known stochastic, continuous, discrete-time, partially-observed, nonlinear, dynamical system with state $\vx \in \mathbb{R}^{d_x}$, observations $\vy \in \mathbb{R}^{d_y}$  and control inputs $\vu \in \mathbb{R}^{d_u}$, operating over a time horizon $T$.
We define the states in upper case to denote the variables over the time horizon, i.e. $\mU = \{\vu_0,\dots,\vu_{T-1}\}$.
The joint distribution (generative model) $p(\mY,\mX,\mU)$ over these variables factorizes into several interpretable distributions: The dynamics $p(\vx_{t+1}|\vx_t,\vu_t)$, observation model $p(\vy_{t}\mid \vx_t,\vu_t)$, and behavior policy $p(\vu_{t}\mid \vx_t)$. 

\subsection{Variational Inference for Latent Variable Models}
\label{sec:vi}
Inference may be described by minimizing the distance between the `true' data distribution $p(\cdot)$ and a parameterized family $q_\vtheta(\cdot)$ \cite{murphy2012machine}.
A popular approach is to minimize the Kullback-Liebler (KL) divergence, e.g. $\min \KL[q_\vtheta\mid \mid p]$ w.r.t. $\vtheta$.
More complex inference tasks can be described by observations $\vy$ influenced by unseen latent variables $\vx$.
Given an observation $\vy^*$, maximizing the likelihood involves integrating over the hidden states, and so is termed the marginal likelihood
$p(\vy^*)= \int p(\vy{=}\vy^*,\vx)d\vx$.
Unfortunately this marginalization is typically intractable in closed-form.
A more useful objective may be obtained by applying a variational approximation of latent state $q_\vtheta(\vx\mid \vy^*)=q_\vtheta(\vx\mid \vy{=}\vy^*)$ to the log marginal likelihood and obtaining a lower bound via Jensen's inequality \cite{murphy2012machine}
\begin{align}
\log \textstyle\int p(\vy^*,\vx)d\vx &= \log \textstyle\int p(\vy^*,\vx)\frac{q_\vtheta(\vx\mid \vy^*)}{q_\vtheta(\vx\mid \vy^*)}d\vx
= \log \E_{\,\vx\sim q_\vtheta(\cdot\mid \vy^*)}\left[\textstyle\frac{p(\vy^*,\vx)}{q_\vtheta(\vx\mid \vy^*)}\right], \\
&\geq \E_{\,\vx\sim q_\vtheta(\cdot\mid \vy^*)}\left[\log\textstyle\frac{p(\vy^*,\vx)}{q_\vtheta(\vx\mid \vy^*)}\right] = \text{-}\KL[q_\vtheta(\vx\mid \vy^*)||p(\vx,\vy^*))],\label{eq:elbo_1}
  \\
&= \E_{\,\vx\sim q_\vtheta(\cdot\mid \vy^*)}[\log p(\vy^*\mid \vx)] - \KL[{q_\vtheta(\vx\mid \vy^*)}\mid\mid p(\vx)],\label{eq:elbo_2}
\end{align}
where equations \ref{eq:elbo_1}, \ref{eq:elbo_2} are variations of the `evidence lower bound objective' (ELBO).
The expectation maximization algorithm (EM) \cite{murphy2012machine}, can be understood via \Eqref{eq:elbo_2} as iteratively estimating the latent states (minimizing the KL term via $q$) in the E step and maximizing the likelihood term in the M step.

\section{Active Inference}
\label{sec:active_inference}
Active Inference frames sensorimotor behaviour as the goal of equilibrium between its current and desired observations, which in practice can be expressed as the minimization of a distance between these two quantities.
This distance is expressed using the KL divergence, resulting in a variational free energy objective as described in Section \ref{sec:vi}.
Curiously, AI is motivated directly by the ELBO, whose negative is referred to in the AI literature as the `free energy' $\mathcal{F}(\cdot)$.
The minimization of this quantity, $\mathcal{F}(\vy^*,\vx,\vu) = \KL[{q_\vtheta(\vx,\vu\mid\vy^*)\mid\mid p(\vx,\vu,\vy^*)}]$, as a model of behavior (i.e. state estimation and control), has been coined the `free energy principle'.

\subsection{Free Energy of the Future}
Despite the ELBO not being temporally restricted, AI delineates a `future' free energy.
This free energy is used to describe the distance between future predicted and desired observations, where $\vu$ is directly represented as a policy $\vu=\pi(\vx)$, so ${\mathcal{F}(\vy^*_t,\vx_t\mid\pi)}$ over the future trajectory is minimized.
In active inference, $\pi$ is commonly restricted to discrete actions or an ensemble of fixed policies, so inferring $p(\pi)$ can be approximated through a softmax $\sigma(\cdot)$ applied to the expected `future' free energies for each policy over $t=[\tau,\dots,T-1]$, with temperature $\gamma$ and prior $p(\pi)$
\begin{align}
	p(\pi\mid\mY^*) \approx  \sigma(\log p(\pi) + \gamma\textstyle\sum_{t=\tau}^{T-1}\mathcal{F}(\vy^*_t, \vx_t, \mid\pi)).
\end{align}
Moreover, for the `past' where $t=[0,\dots,\tau-1]$, minimizing $\mathcal{F}(\cdot)$ amounts for state estimation of $\vx$ given $\vy$.
Another consideration is whether the dynamic and observation models are known or unknown.
In this work we assume they are given, but AI can also include estimating these models from data.

\subsection{Active Inference in Practice}
\label{sec:practice}
Initial AI work was restricted to discrete domains and evaluated on simple gridworld environments \cite{friston2017active,friston2009reinforcement}.
Later work on continuous state spaces use various black-box approaches such as cross-entropy \cite{tschantz2020reinforcement}, evolutionary strategies \cite{ueltzhoffer2018deep}, and policy gradient \cite{millidge2019deep} to infer $\pi$.
A model-based method was achieved by using stochastic VI on expert data \cite{8616685}.
Connections between AI and CaI, performing inference via message passing, have been previously discussed \cite{van2019application,factor_graph}.
AI has been applied to real robots for kinematic planning, performing gradient descent on the free energy using the Laplace approximation every timestep \cite{oliver2019active}.
Despite these various approaches, AI has yet to demonstrate the sophisticated control achieved by advanced optimal methods, such as differential dynamic programming \cite{tassa2012synthesis}.

\section{Control as Inference}
\label{sec:cai}
From its origins in probabilistic control design \cite{karny1996towards}, defining a state $\vz\in \mathbb{R}^{d_z}$
to describe the desired system trajectory\footnote{while $\vz$ could be defined from $[\vx,\vu]\tran$, it could also include a transformation, e.g. applying kinematics to joint space-based control for a  cartesian space objective.}
$p(\mZ)$, optimal control can be expressed as finding the state-action distribution that minimizes the distance for a generative model parameterized by $\vtheta$, which can be framed as a likelihood objective \cite{murphy2012machine}
\begin{align}
	\min\; &\KL[p(\mZ) \mid \mid  q_\vtheta(\mZ)]
	\equiv 
	\max\; \E_{\mZ\sim p(\cdot)}[\log \textstyle\int q_\vtheta(\mZ,\mX,\mU)d\mX d\mU]. \label{eq:cai}
\end{align}
When $p(\mZ)$ simply describes a desired state $\vz_t^*$, so
${p(\vz_t)=\delta(\vz_t-\vz_t^*)}$,
and the latent state-action trajectory is approximated by
$q_\vphi(\mX,\mU)$, the objective (\Eqref{eq:cai}) can be expressed as an ELBO where the `data' is $\mZ^*$ 
\begin{align}
	\max \E_{\,\mX,\mU\sim q_\vphi(\cdot\mid \mZ^*)}[\log q_\vtheta(\mZ^*\mid\mX,\mU)]{-} \KL[q_\vphi(\mX,\mU\mid\mZ^*) \mid q_\vtheta(\mX,\mU)],
\end{align}
where $\vphi$ captures the latent state parameterization and $\vtheta$ defines the remaining terms, i.e. the priors on the system parameters and latent states.
This objective can be optimized using EM, estimating the latent state-action trajectory $\vphi$ in the E step and optimizing the remaining unknowns $\vtheta$ in the M step.
By exploiting the temporal structure, $q_\vphi(\mX,\mU\mid\mZ^*)$ can be inferred efficiently in the E step by factorizing the joint distribution (\Eqref{eq:i2c_joint}) and applying Bayes rule recursively
\begin{align}
&q_\vphi(\mZ^*,\mX,\mU){=}q_\vphi(\vx_0)
\textstyle\prod_{t=0}^{T-1}{q_\vphi(\vx_{t+1}|\vx_{t}, \vu_{t})}
\textstyle\prod_{t=0}^{T} q_\vphi(\vz^*_{t}|\vx_{t}, \vu_{t}) q_\vphi(\vu_{t}|\vx_{t}),\label{eq:i2c_joint}
\end{align}
\vspace{-2.7em}
\begin{align}
q_\vphi(\vx_{t},\vu_{t}\mid\vz^*_{0:t}) &\propto q_\vphi(\vz^*_{t}\mid\vx_{t},\vu_{t})\, q_\vphi(\vx_{t},\vu_{t}\mid\vz^*_{0:t-1}),\label{eq:filter}\\
q_\vphi(\vx_{t},\vu_{t}\mid\vz^*_{0:T}) &\propto q_\vphi(\vx_{t},\vu_{t}\mid\vx_{t+1})\, q_\vphi(\vx_{t+1}\mid\vz^*_{0:T}).\label{eq:smooth}
\end{align}
Equations \ref{eq:filter}, \ref{eq:smooth} are commonly known as Bayesian filtering and smoothing \cite{10.5555/2534502}.
The key distinction of this framework from state estimation is the handling of $\vu$ during the forward pass, as $q_\vphi(\vx_t,\vu_t){=}q_\vphi(\vu_t\mid\vx_t)q_\vphi(\vx_t)$, control is incorporated into the inference.
We can demonstrate this in closed-form with linear Gaussian inference and linear quadratic optimal control.
 
\subsection{Linear Gaussian Inference \& Linear Quadratic Control}
\label{sec:linear_gaussian}
While the formulation above is intentionally abstract, it can be grounded clearly by unifying linear Gaussian dynamical system inference (LGDS, i.e. Kalman filtering and smoothing) and linear quadratic Gaussian (LQG) optimal control \cite{toussaint2009robot}.
While both cases have linear dynamical systems, here LQG is fully-observed\footnote{Confusingly, LQG can refer to both Gaussian disturbance and/or observation noise. While all varieties share the same optimal solution as LQR, the observation noise case results in a partially observed system and therefore requires state estimation. \itwoc{} is motivated by the LQR solution and therefore does not consider observation noise, but it would be straightforward to integrate.} and has a quadratic control cost, while the LGDS is partially observed and has a quadratic log-likelihood due to the Gaussian additive uncertainties. 
These two domains can be unified by viewing the quadratic control cost function as an Gaussian observation likelihood. For example, given $\vz_t = \vx_t + \vxi, \vxi\sim\N(\vzero,\mSigma)\text{ and }\vz_t^* = \vzero\;\forall\; t$,
\begin{align}
	\log q_\vtheta(\vz_t^*|\vx_t,\vu_t) &=
	\text{-}\textstyle\frac{1}{2}(d_z\log 2\pi + \log|\mSigma| + \vx_t\tran\mSigma\inv\vx_t) = \alpha \vx_t\tran\mQ\vx_t + \beta
\end{align}
where $(\alpha,\beta)$ represents the affine transformation mapping the quadratic control cost $\vx\tran\mQ\vx$ to the Gaussian likelihood.
As convex objectives are invariant to affine transforms, this mapping preserves the control problem while translating it into an inference one.
The key unknown here is $\alpha$, which incorporates $\mQ$ into the additive uncertainty $\vxi$, $\mSigma=\alpha\mQ\inv$.
Moreover, inference is performed by using message passing \cite{loeliger2007factor} in the E step to estimate $\mX$ and $\mU$, while $\alpha$ is optimized in the M step.
This view scales naturally to not just the typical LQG cost $\vx\tran\mQ\vx + \vu\tran\mR\vu$, but also nonlinear mappings to $\vz$ by using approximate inference.
While the classic LQG result includes the backward Ricatti equations and an optimal linear control law, the inference setting derives direct parallels to the backward pass during smoothing \cite{toussaint2009robot} and the linear conditional distribution of the Gaussian, $q_\vtheta(\vu_t\mid \vx_t){=}\N(\mK_t\vx_t + \vk_t, \mSigma_{k_t})$ \cite{hoffmann2017linear} respectively.
As the conditional distribution is linear, updating the prior joint density $p(\vx_t, \vu_t)$ in the forward pass with updated state estimate $\vx_{t}'$ corresponds to linear feedback control w.r.t. the prior
\begin{align}
p(\vu'_t) &=
\int p(\vu_t|\vx_t{=}\vx_t')p(\vx'_t)d\vx'_t,\\
\vmu_{\vu_t'} &= \vmu_{\vu_t} + \mK_t(\vmu_{\vx_t}-\vmu_{\vx'_t}),\\
\mSigma_{{\vu\vu}'_t} &= \mSigma_{{\vu\vu}_t} - \mSigma_{{\vu\vx}_t}\mSigma_{{\vx\vx}_t}\inv\mSigma_{{\vx\vu}_t}\tran
+ \mK_t \mSigma_{{\vx\vx}'_t} \mK_t\tran,\\
\mK_t &= \mSigma_{{\vu\vx}_t}\mSigma_{{\vx\vx}_t}\inv\label{eq:gain}.
\end{align}
From \Eqref{eq:gain}, it is evident that the strength of the feedback control depends on both the certainty in the state and the correlation between the optimal state and action. 

The general EM algorithm for obtaining $q_\vtheta(\vx,\vu)$ from $p(\mZ)$ is referred to as input inference for control (\itwoc) \cite{i2corl} due to its equivalence with input estimation.
Note that for linear Gaussian EM, the ELBO is tight as the variational distribution is the exact posterior.
For nonlinear filtering and smoothing, mature approximate inference methods such as Taylor approximations, quadrature and sequential Monte Carlo may be used for efficient and accurate computation~\cite{10.5555/2534502}.
\newline\newline\noindent
Another aspect to draw attention to is the inclusion of $\vz$ compared to alternative CaI formulations, which frame optimality as the probability for some discrete variable $\vo$, $p(\vo{=}1\mid\vx,\vu)$ \cite{levine2018reinforcement}.
Previous discussion on CaI vs AI have framed this discrete variable as an important distinction.
However, it is merely a generalization to allow for a general cost function $C(\cdot)$ to be framed as a log-likelihood, i.e. $p(\vo{=}1\mid\vx,\vu) \propto \exp(-\alpha C(\vx,\vu))$.
For the typical state-action cost functions that are a distance metric in some transformed space, the key consideration is the choice of observation space $\vz$ and corresponding exponential density.

\begin{figure}[!b]
	\vspace{-1em}
	\begin{tikzpicture}

\definecolor{color0}{rgb}{0.75,0.75,0}
\definecolor{color1}{rgb}{0.75,0,0.75}
\definecolor{color2}{rgb}{0,0.75,0.75}

\begin{axis}[
hide axis,
width=2cm,
height=5cm,
xmin=10, xmax=50,
ymin=0, ymax=1.0,
legend pos=north east,
legend columns=5,
legend style={
/tikz/every even column/.append style={column sep=1cm},
draw=none,
inner ysep=4pt
}]
]

\addlegendimage{color0, mark=*, mark size=2}
\addlegendentry{Prior};
\addlegendimage{color2, mark=*, mark size=2}
\addlegendentry{Posterior};
\addlegendimage{black, dashed, mark=x, mark size=2}
\addlegendentry{$y^*$};
\addlegendimage{black, dashed, mark=*, mark size=2}
\addlegendentry{$z^*$};
\addlegendimage{red, mark=x, mark size=2, mark options={solid}, only marks}
\addlegendentry{\textsc{lqr}};

\end{axis}

\end{tikzpicture}
\vspace*{\fill}
	\input{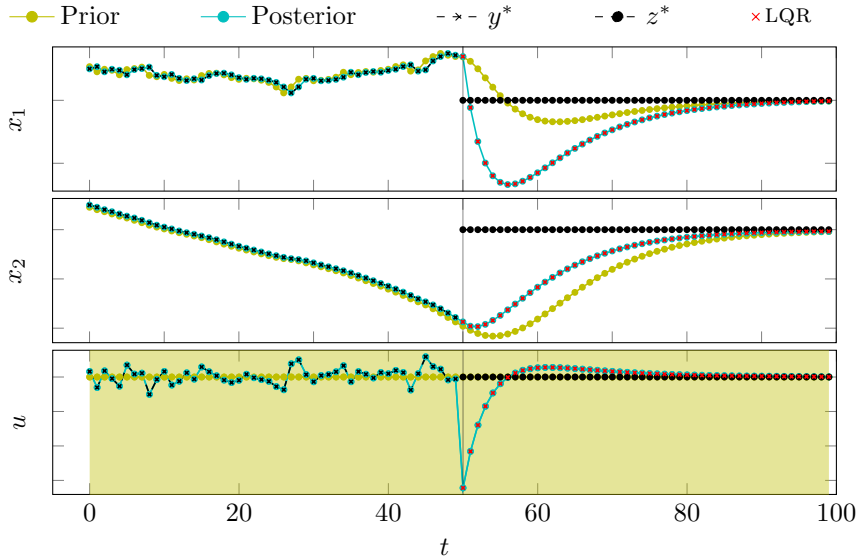}
	\vspace{-1em}
	\caption{
		Linear Gaussian \itwoc{} performing state estimation and control following Section \ref{sec:unify}, with state $\vx{=}[x_1, x_2]\tran$, action $u$ and $[\vx, u]\tran$ as the observation space. With $\tau=50$, for $t<\tau$ $\itwoc$ performs state estimation under random controls. For $t\geq\tau$, $\itwoc$ switches to optimal control. This example is in the low noise setting, with a large prior on $u$, to illustrate that \itwoc{} returns the LQR solution for the same initial state and planning horizon.
	}
	\label{fig:traj}
	
\end{figure}

\section{The Unifying View: Control of the Observations}
\label{sec:unify}
A key distinction to the AI and CaI formulations described above is that, while AI combines state estimation and control with a unified objective, CaI focuses on trajectory optimization.
However, this need not be the case.
In a similar fashion to the partially-observed case of LQG, CaI also naturally incorporates observations \cite{toussaint2008hierarchical}.  
As Section \ref{sec:cai} describes \itwoc through a general Bayesian dynamical system, the formulation can be readily adapted to include inference using past measurements.
Moreover, as \itwoc frames the control objective as an observation likelihood, when $\vz$ and $\vy$ are the same transform of $\vx$ and $\vu$, the objective can also be unified and directly compared to active inference.
For `measurements' $\mY^* = \{\vy^*_0,\dots,\vy_{\tau\minus 1}^*,\vz^*_{\tau},\dots,\vz_{T-1}^*\}$, following \Eqref{eq:cai} using the $\mathcal{F}(\cdot)$ notation
\begin{align}
\min\; &\KL[p(\mY)||q_\vtheta(\mY)]{=}
\min\;
\underbrace{
\sum_{t=0}^{\tau\minus 1}
\mathcal{F}_\vpsi(\vy^*_t,\vx_t,\vu_t)}_{\text{state estimation}}
{+}
\underbrace{
\sum_{t=\tau}^{T\minus 1}
\mathcal{F}_\vpsi(\vz^*_t,\vx_t,\vu_t),}_{\text{optimal control}}
\label{eq:unify}
\end{align}
where $\vpsi=\{\vtheta,\vphi\}$.
Here, $p(\vy_t){=}\delta(\vy_t-\vy_t^*)$ now also describes the empirical density of past measurements $\vy_{<\tau}^*$ .
The crucial detail for this representation is that the observation model $q_\vtheta(\vy_t\mid\vx_t,\vu_t,t)$ is now time dependent, switching from estimation to control at $t=\tau$.
For the Gaussian example in Section \ref{sec:linear_gaussian}, $\mSigma_{<\tau}$ is the measurement noise and $\mSigma_{\geq\tau}\inv{=}\alpha\mQ$.
A benefit of this view is that the computation of active inference can now be easily compared to the classic results of Kalman filtering and LQG (Fig. \ref{fig:traj}), and also scaled to nonlinear tasks through approximate inference.
Moreover, obtaining the policy $\pi(\cdot)$ using the joint distribution $q_\vtheta(\vx_t,\vu_t)$ is arguably a more informed approach compared to direct policy search on an arbitrary policy class.

\section{Conclusion}
We have derived an equivalent formulation to active inference by considering partially-observed, inference-based optimal control, which has a principled derivation and is well-suited for approximate inference.
While we have delineated state estimation as operating on past measurement and control as planning future actions (\Eqref{eq:unify}), both AI and \itwoc{} demonstrate the duality between estimation and control due to the mathematical similarity when both are treated probabilistically.   
We hope the inclusion of the CaI literature enables a greater theoretical understanding of AI and more effective implementations through approximate inference.

\bibliographystyle{splncs04}
{\footnotesize\bibliography{bibliography}}

\end{document}